# The state-of-the-art review on resource allocation problem using artificial intelligence methods on various computing paradigms

Javad Hassannataj Joloudari[1,*], Sanaz Mojrian[2], Hamid Saadatfar[1], Issa Nodehi[3], , Fatemeh Fazl[4], Sahar Khanjani shirkharkolaie[2], Roohallah Alizadehsani[5], H M Dipu Kabir[5], Ru-San Tan[6], U Rajendra Acharya[6,7,8]

[1]Department of Computer Engineering, Faculty of Engineering, University of Birjand, Birjand 9717434765, Iran
[2]Department of Information Technology, Mazandaran University of Science and Technology, Babol, Iran
[3]Department of Computer Engineering, Qom University of Technology, Qom, Iran
[4]Department of Electronic Engineering, Faculty of Electrical and Computer Engineering, University of Birjand, Birjand, Iran
[5]Institute for Intelligent Systems Research and Innovation, Deakin University, Geelong, VIC 3216, Australia
[6]School of Science and Technology, Singapore University of Social Sciences, Singapore
[7]Ngee Ann Polytechnic, Singapore, 599489, Singapore
[8]Dept. of Biomedical Informatics and Medical Engineering, Asia University, Taichung, Taiwan

[*]Correspondence: javad.hassannataj@birjand.ac.ir

**Abstract**

With the increasing growth of information through smart devices, increasing the quality level of human life requires various computational paradigms presentation including the Internet of Things, fog, and cloud. Between these three paradigms, the cloud computing paradigm as an emerging technology adds cloud layer services to the edge of the network so that resource allocation operations occur close to the end-user to reduce resource processing time and network traffic overhead. Hence, the resource allocation problem for its providers in terms of presenting a suitable platform, by using computational paradigms is considered a challenge. In general, resource allocation approaches are divided into two methods, including auction-based methods (goal, increase profits for service providers-increase user satisfaction and usability) and optimization-based methods (energy, cost, network exploitation, Runtime, reduction of time delay). In this paper, according to the latest scientific achievements, a comprehensive literature study (CLS) on artificial intelligence methods based on resource allocation optimization without considering auction-based methods in various computing environments are provided such as cloud computing, Vehicular Fog Computing, wireless, IoT, vehicular networks, 5G networks, vehicular cloud architecture, machine-to-machine communication (M2M), Train-to-Train(T2T) communication network, Peer-to-Peer (P2P) network. Since deep learning methods based on artificial intelligence are used as the most important methods in resource allocation problems; Therefore, in this paper, resource allocation approaches based on deep learning are also used in the mentioned computational environments such as deep reinforcement learning, Q-learning technique, reinforcement learning, online learning, and also Classical learning methods such as Bayesian learning, Cummins clustering, Markov decision process.

## 1  Introduction

With the rapid growth of the use of network-based methods, several computational platforms are created. By creating these platforms in a scalable and executable system, a technology called the Internet of Things (IoT) is formed .

The advantages of this technology are Common data transfer of platforms and infrastructures, integration and synchronization of systems in a distributed system underlying IoT.

Cyber-physical systems are an example of IoT technology in the interaction between humans and objects via the Internet which has been used in recent years in various industries including health, transportation, Smart houses [1-5].

In the IoT environment, objects can be hardware and software applications that their goal is to meet users' expectations by offering them suggestions [6]. Connected devices can always be easily displayed as objects on the Internet platform. These devices provide and allocate resources with the highest quality to the users of a system. Given that, today we are facing the expansion of smart devices, there is no integrated and compatible mechanism with such devices to provide appropriate and complete resource allocation. The most common alternative solution for allocating resources to different users in the IoT environment is to use smart agents and tools. The main purpose of these smart agents is high performance to obviate the needs of a user. So that the high performance should include high-level features of a system such as power consumption, response time, security level. These smart agents and tools need to create a variable cost mapping table. Updating this mapping table for smart agents is a challenging problem, while a smart agent must acceptably optimize the cost of using resources. Therefore, it is necessary to use methods based on computational intelligence, such as supervised and reinforced learning methods called machine learning methods [7, 8]. Also, deep reinforcement learning methods can be used in the resource optimization process in various computing paradigms [9]. In recent years, cloud computing technology is known as the most popular computing environment on the Internet. The conventional cloud computing model is shown in Figure 1.

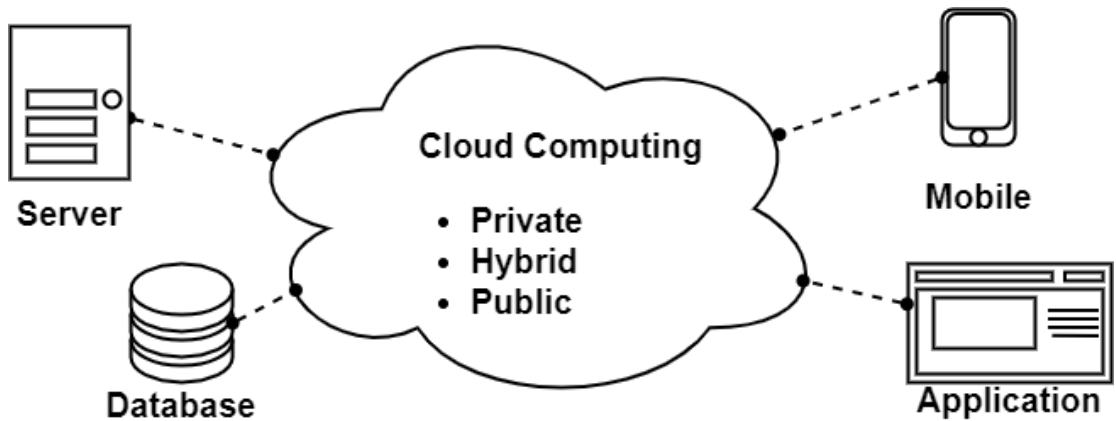

**Figure 1.** conventional cloud computing model [10-13].

According to Figure 1, cloud computing has three types of clouds: public (describing the traditional and classic model of cloud computing like Google App Engine), private (describing a cloud computing infrastructure through an organization for internal use of that organization, such as Amazon virtual cloud) and hybrid (a combination of two models of public cloud and private cloud). A pre-contract is established between cloud service providers and users to consider the type of cloud for users. Generally, users' required applications are stored in a database on a cloud server, then the decision about the type of cloud is made by the administrator of a cloud system. Also, a series of applications can be transferred through smart devices (mobile) to the server of a cloud system according to the selected cloud type. In addition, this technology is managed by resource providers according to the needs of users with Computational resource virtualization such as memory, communication bandwidth, disk, CPU, and a variety of software and platforms. Therefore, cloud computing has three types of services, including infrastructure as a service (Iaas: Linode, Rackspace, cisco metapod), platform as a service (pass: windows azure, Heroku, Google App Engine), and software as a service (such as Saas: google apps, salesforce, cisco webEx). Since the allocation of resources in cloud data centers by virtual machines is done at the request of users. Therefore, the Effective compatibility of virtual machines is very important in empowering the cloud computing pattern. To achieve this feature, it is essential to use effective strategies for resource allocation and virtual machine management [14]. Therefore, to solve the resource allocation problem in cloud-based systems, automatic decision-making methods should be used. In recent years, machine learning-based methods and deep learning have been used for cloud-based systems. These methods are very suitable for increasing the power and performance of a system. They also do not require initial modeling of state transition and workload. In particular, agents based on the reinforcement learning method can be trained to decide automatically on the optimal allocation of resources. And also can control the activities of a system underlying cloud computing compatibly and online [14]. One of the main disadvantages of using cloud computing, which poses major challenges is such as the long-distance data transfer and time delay which makes the quality of services largely not guaranteed. More recently, an emerging computing technology called cloud computing has been placed between IoT and cloud levels, with the goal of better resource management, proper data preprocessing, short-time delay for data transfer, and network graphics overhead Reducing. According to Cisco, fog computing technology Extends cloud services to the edge of the network by connecting the cloud layer to the IoT layer and occurs close to the end-user which reduces data processing time and network traffic overhead. In general, the fog layer is another layer of distributed networks that has a close connection to cloud computing and the Internet of Things which guarantee service quality for devices and other applications that require interactions in real-time [15]. Advantages of fog calculations include reduction of time delay, very low jitter, location of service on the edge of the local network, one-step customer-server communication, definable security, a large number of server nodes, and support for real-time or real-time interactions. The most basic entity in fog is the fog node, which facilitates the execution of IoT applications [16]. Any device with a network connection, computing, and storage can be a node. For example, switches, routers, hubs, industrial controllers, and surveillance cameras are considered nodes in the fog. A new paradigm has been introduced in computational systems underlying edge computing [17]. This paradigm is partial or a subset of fog calculations. In a way, the idea of fog calculations is the method of information processing from where it is generated to where it is stored while edge calculations are only attributed to processed information close to where it was created. Fog calculations not only cover edge processing but also the required network connections to bring that data from the edge to its endpoint. A three-layer framework for computing environments is shown in Figure 2, such as the Internet of Things, fog calculation and cloud computing.

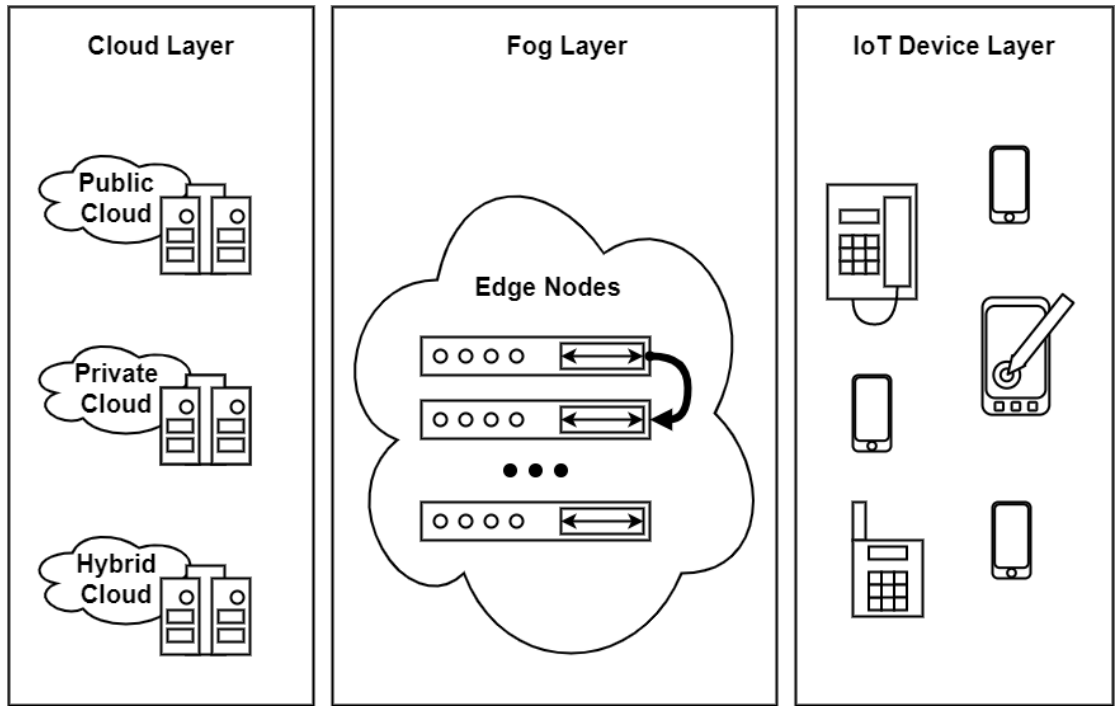

**Figure 2.** Multilayer framework for computational environments [15, 18].

According to Figure 2, considered layers based on service allocation include: IoT application layers, fog computing layer, and cloud layer. It should be noted that the edge computing paradigm is a subset of the fog computing layer. IoT applications consist of a large number of services requested by users of a system that must be responded to in real-time by the fog layer and the cloud layer. For this purpose, with the development of computing resources to edge computing for cloud computing, resource allocation on LAN has a great advantage for mobile users who are close to the network edge in terms of time delay and short distance. The main purpose of edge calculations is to minimize data transfer delays from the Internet of Things layer to the cloud layer by edge calculations on fog calculations and increase service quality [19]. To the best of our knowledge, a comprehensive literature study (CLS) on resource allocation problems using artificial intelligence methods is provided for this paper including machine learning and deep learning in smart computing environments. As yet, no comprehensive review of artificial intelligence methods has been conducted for resource allocation in computing environments. The examined computing environments in this paper include cloud computing, Vehicular Fog Computing, wireless, Internet-of-Things (IoT) system, vehicular networks, 5G networks, vehicular cloud computing, machine-to-machine communication (M2M), Train-to-Train(T2T) communication network, and Peer-to-Peer (P2P) network, Mobile Cloud Computing (MCC), Cellular and wireless IoT networks.

A summary of the innovations of this review article is as follows.
- Comprehensive Literature Study (CLS) on machine learning and deep learning-based methods for resource allocation problem in emerging computing environments

- Comparison between artificial intelligence methods concerning resource allocation problem
- Present open issues and future research challenges in resource allocation on multilayer computing environments

The rest of this article is as follows:

In the second part, a comprehensive literature study (CLS) will be presented on resource allocation problems, review articles conducted up to 2020, and also two taxonomies underlying machine learning and deep learning methods. The comparison between artificial intelligence methods in resource allocation will be described in the third section. Also, in section 4, open research challenges will be stated. Finally, conclusions and future research work are described in section 5.

**Notation:**

Table 1 describes the abbreviations and acronyms related to intelligent computing environments in this study.

**Table 1.** List of abbreviations related to intelligent computing environments.

| Abbreviations | Meaning |
| --- | --- |
| IOT | Internet of Thing |
| CC | Cloud Computing |
| FC | Fog Computing |
| MEG | Mobile Edge Computing |
| P2PN | Peer-to-Peer Network |
| 5GN | 5G Network |
| VCC | Vehicular Cloud Computing |
| M2MN | Machine-to-Machine Network |
| VFC | Vehicular Fog Computing |
| WN | Wireless Network |
| VN | Vehicular Network |
| MCC | Mobile Cloud Computing |
| T2TN | Train-to-Train Network |
| QL-KM | Q- learning, K-means |
| CLS | comprehensive literature study |
| MDP | Markov Decision Process |
| P2P | Peer-to-Peer |
| M2M | machine-to-machine |
| DRL | Deep Reinforcement Learning |

## 2 Comprehensive literature study for RA problem

The Comprehensive Literary Study (CLS) for the resource allocation problem has been performed in previous studies by using computational intelligence methods including machine learning and deep learning methods. In this article, we express in detail the works of others in resource allocation. This section includes three subsections 2.1, 2.2, and 2.3 that we express in, 2.1 section:

the review of articles conducted up to 2020, in 2.2 section, the first taxonomy related to ML methods, and also in 2.3subsection, the second taxonomy related to DL methods.

## 2.1 The survey articles related to resource allocation

We describe the review works which has been done by researchers for resource allocation problem until 2020.

In a study by Yousefzai et al. [20], the resource allocation problem in the cloud computing environment has been investigated. They examined the schemes based on cloud computing resources by using effective features such as optimization goals, optimization methods, design approaches, and useful functions in their study.

Atman and Nayan [21] reviewed the solutions based on reinforcement and heuristic learning to use a dynamic and adaptive allocation resource. They categorized this method for public safety communications on fifth-generation mobile networks and the results of the implementation of these methods were examined in other articles. In these reviews, deep learning methods had faster and more convergence that is more accurate. As a result, the Evolutionary RL / GA algorithm suggested in [22] works more than 90% better than the naïve strategy in the long-term use of the network according to the global optimum. For example, the RL method is shown in Figure 3. Also, the comparison results of the Value-based RL / Q-learning algorithm suggested in [23] show that approximate dynamic programming has a faster convergence rate and maximum sum rate.

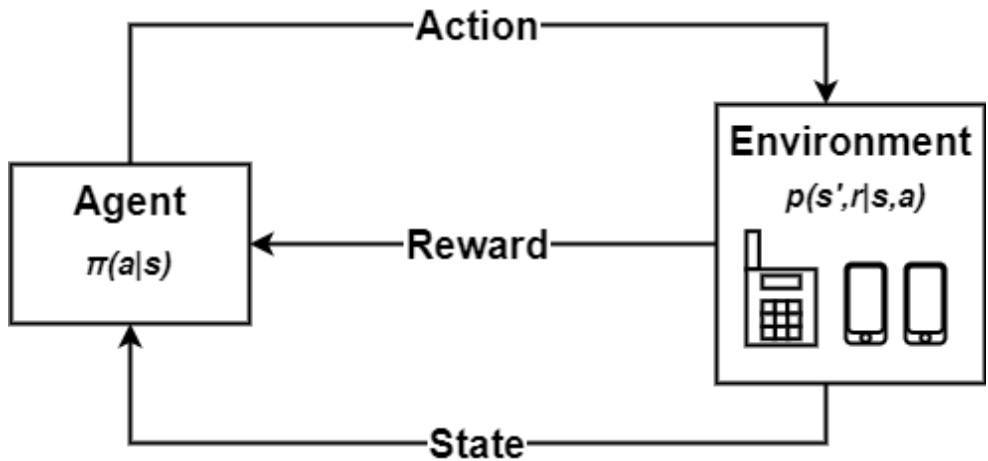

**Figure 3.** The agent-environment interaction in RL.

Based on Figure 3, at each distinct time stage $t$, the agent observes representations of the environment state $S_t$ from state space $S$, and then chooses an action $A_t$ from the set of actions $A$. After the action, the agent accepts a reward $R_{t+1}$, then a new state will be established in the environment $S_{t+1}$, with the probability of $p(s',r|s,a)$. In RL, a policy $\pi(a|s)$ will be used for decision making, that is a mapping of states in $S$ to probabilities of choosing one action in $A$. The learning is performed to discover an optimal policy $\pi^*$ that get the most out of the probable growing rewards from one primary state $s$.

In another study by Qobaei et al. [24], six categories of resource management were examined, including application placement, resource scheduling, task loading, load balancing, resource allocation, and resource provisioning for the computing environment, So that they considered two approaches based on auction and optimization to allocate the resource allocation problem.

The reviewed articles by Hameed et al. [25], Beloglazov et al. [26], Shuja et al. [27], are based on effective energy for the resource allocation problem.

Aceto et al. [28] focused on resource monitoring in the cloud computing environment.

A review article conducted by Jennings and Stadler [29], developed as a framework for resource management in the cloud computing environment.

Goyal and Dadizadeh [30] pay attention to the Implementation details of parallel processing frameworks, including Google MapReduce and Microsoft's Dryad.

Hussain et al. [31] presented the work process for commercial cloud computing service providers and open-source deployment solutions.

Huang et al. [32] examined the dynamic resource allocation problem. They also studied task scheduling strategies. Their examination shows how a system with a SaaS-based cloud computing service operates under existing infrastructure.

Ahmed et al. in [33] and [34] studied the Virtual machine migration optimization features underlying cloud data center service operators.

A review study conducted by Vinothina et al. [35] analyzed the classification of types of strategies and challenges related to resource allocation and their effects on the cloud computing system. They focused specifically on CPU and memory resources regarding the strategies implemented for resource allocation.

In a study by Anuradha and Sumathi [36], resource allocation techniques and strategies in the cloud computing environment were examined. They made a comparison between merits and demerits techniques, and their examined strategy consists of prediction algorithms for resource requirements and resource allocation algorithms. As a result, they set to identify efficient resource allocation strategies with effective use of resources in a cloud computing environment with limited resources.

Mohamaddiah et al. [37] researched in the field of resource management, In particular resource allocation and resource monitoring strategies, and also problem-solving approaches of resource allocation in the cloud environment.

Rama Mohan and Baburaj [38] provided strategies for resource allocation and their applications in the cloud. In their study, the issue of resource allocation adapted in the cloud environment based on various Dynamic proportions was explained in detail.

Casta͠neda et al. [39] provided an overall overview of a variety of techniques to achieve common optimization tasks downlink related to Multi-User Multiple-Input communication systems.

In another study, Manvi and Shyam [40] examined the management methods of resource management approaches such as resource provisioning, resource allocation, resource matching, and resource mapping, and also they provided an overall overview of methods for IaaS in the cloud computing environment.

Su et al. [41] examined the techniques and models of resource allocation algorithms in 5G network slicing. They expressed ideas about software-defined networking and network functions virtualization and also their tasks in network slicing. In addition, the management and orchestration architecture of the network slice was also presented, which is a fundamental framework for resource allocation algorithms.

To the best of our knowledge and the latest scientific achievements, no review article has been conducted until now for resource allocation optimization problems by using artificial intelligence-based methods such as machine learning and deep learning in a variety of computing environments. In the following, we will examine the methods based on machine learning and deep learning in subsections 2.2 and 2.3, respectively.

## 2.2 The first Taxonomy related to ML methods

Taxonomy related to ML methods for resource allocation problems in different computational environments is shown in Figure 4.

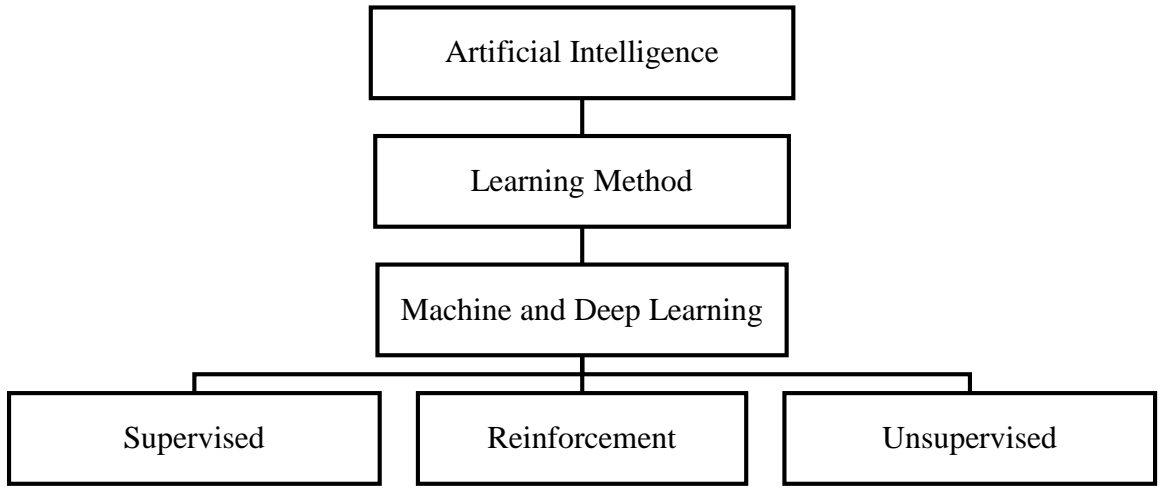

**Figure 4.** Taxonomy related to machine learning methods for resource allocation problems in different computational environments.

According to figure 4, ML methods for RA are divided into three portions such as supervised learning, reinforcement learning, and unsupervised learning.

### 2.2.1 Supervised Learning of ML methods for RA

Shi et al. [42] provided the Markov Decision Process (MDP) method and Bayesian learning to optimize the cost of dynamic resources allocation in the cloud computing environment for the components of Network functions virtualization [43]. The results of their experiments show that the MDP method helps to allocate cloud resources in the components of the Network functions virtualization. And also, the Bayesian learning method predicts high reliability in using resources to increase the performance of cloud resource management based on virtualization of network functions in the future. In addition, their proposed method was better than the greedy methods such as dynamic scaling, cost modeling, and VM placement in terms of the total cost of cloud resource allocation time.

Rohmer et al. [8] used a resource allocation framework underlying learning called LB-RA (for P2P streaming systems). In their study, they used collected real data conducted by Xu Cheng et al. [44]. According to the LB-RA approach, they proposed the Bayesian method for switching between resource allocations strategies in P2P systems. As a result, the proposed approach (LB-RA) shows the best performance to the (Lowest Popularity Score, Lowest Critical Score, Highest

Uplink First, Greedy) methods in terms of Mean rejection rate with 9.2%, Max rejection rate with 55.2%, Mean entropy value with 6.20 Entropy standard deviation with 0.87 shows the best performance.

### 2.2.2 Reinforcement Learning of ML methods for RA

Gai and Qiu [4], by focusing on the issue of dynamic resource allocation, using the Quality of Experience Level (QoE) metric with two reinforcement-based algorithms, including Reinforcement Learning-based Mapping Table (RLMT) to updating / maintaining The cost table and the Reinforcement Learning-based Resource Algorithm(RLRA) used the IoT environment as a high-level IoT to achieve the quality of experience for achieving Smart Content-Centric for Internet-of-Things (SCCS-IoT) in a cyber-physical system. In general, their findings according to Reinforcement Learning-based algorithm show that the number of computational nodes has a significant effect on training time. As the number of input tasks increases, training time increases, and also grouping computational nodes with similar capabilities can shorten the training period.

AlQerm and Shihada [7] proposed a participatory online learning algorithm for optimal allocation with power and modulation adaptation capability in 5G systems. So that, they solved the problem of interference, including cross-tier interference and co-tier interference in their proposal. They showed that the proposed plan is better than the Down-SA [45], Joint-RALA [46], and Matching-RM [47] schemes and has improved significantly in the field of throughput, spectral efficiency, fairness, and outage ratio for different underlay edge transitions compared to other plans.

Hamidreza Arkian et al. [48] used the Improved COHORT architecture [49] based on clustering for resources management, increase efficiency, stability, and reliability of vehicular cloud architecture. So that. They developed this architecture with the Q-learning algorithm and three queuing strategies underlying the cloud architecture. In their study, they also used the method based on introduced fuzzy logic to select the eclipse. In their experiment, they compared the proposed COHORT clustering plan with two plans based on user-oriented fuzzy logic-based clustering Scheme [50] and Lowest-ID [51]. The results of the experiments show that by changing the maximum speed, from 60 to 120 km/h, the cluster head duration for the proposed COHORT scheme decrease about 15%, while for both Lowest-ID and user-oriented techniques, the value is much lower. In addition, they compared the CROWN [52] and COHORT architectures in terms of service discovery delay, so that with the increase in the number and density of vehicles, the service discovery delay for COHORT architecture is much less than CROWN, and also by comparing these architectures in terms of Service consumption delay, the results show, as the number of vehicles increases, service consumption delay decreases dramatically for the COHORT architecture relative to the CROWN.

Hussain et al. [53] used the Q-learning algorithm to slot assignment in the machine-to-machine communication network and the k-means clustering algorithm to overcome congestion. The results of the experiments show that The Q-learning algorithm increases the probability of slot assignment by more than five times compared to Ethernet slot assignment protocols such as ALOHA, slotted Aloha, and channel-based allocation reduces the learning rate and increases the probability of convergence.

Salahuddin et al. [19] proposed two methods, the Markov decision process (MDP) and greedy heuristics [54] to minimize overhead in the Vehicular Cloud environment. Their results show that the MDP method has better performance in terms of Long- term benefit and minimizing overhead in terms of resource provisioning. In addition, a comparison between the MDP and the greedy heuristic method shows that the two methods lead to the same configuration choice, and also a comparison of the output of the MDP and myopic heuristic methods indicates that MDP selects the configuration with the lowest overhead in the long run, and in the worst case, MDP works as well as a myopic heuristic.

## 2.3 The second Taxonomy related to DL methods

Taxonomy related to DL methods for resource allocation problems in various computational paradigms is shown in Figure 5.

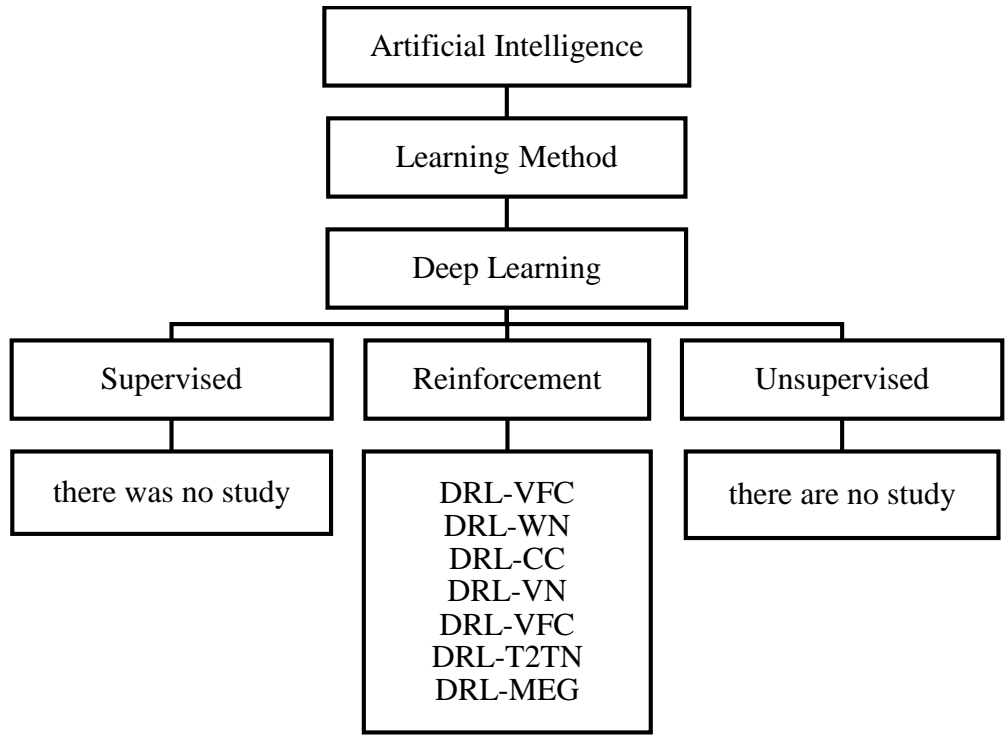

**Figure 5.** Taxonomy related to DL methods for the resource allocation problem in various computational paradigms.

Based on figure 5, DL methods for RA are divided into three portions such as supervised learning, reinforcement learning, and unsupervised learning.

### 2.3.1 Reinforcement Learning of DL methods for RA

Karthiban and Raj [13] used a deep reinforcement learning algorithm for fair resource allocation to achieve a better resource allocation model in the cloud computing environment. They compared their proposed algorithm with FIFO and greedy methods. The results of the experiment, in terms of criteria of the average response time and average waiting time, with an increasing number of requests, compared to FIFO and greedy methods, show that the proposed deep reinforcement learning model works better and has an optimal solution, and ensures that QoS is guaranteed in exchange for providing optimal planning.

Chen et al. [55] have provided a deep reinforcement learning method in time-limited resource allocation to reduce overall delay, underlying the Perception reaction time (PRT) criteria in the field of Vehicular applications for the fog computing environment under Information-Centric Network-Internet of Vehicular (ICN-IOV). In this paper, resource allocation to vehicles was done by the Markov decision process and also PRT criterion of vehicle safety / non-safety applications was minimized by applying a deep reinforcement learning algorithm. This algorithm has better performance than Q-learning, location greedy and resource greedy algorithms during convergence, and also the architecture based on the Vehicular fog computing environment with PRT criterion is more stable than architectures such as No fog, No ICN, and No fog r ICN so that the PRT criterion is reduced by about 70%.

Ye et al. [56] have provided a deep reinforcement learning method for decentralized resource allocation in the field of Vehicle-to-vehicle (V2V) communications. This method can be used in unicast and multicast vehicle communications. In their study, resource allocation based on deep reinforcement learning has higher profit in V2V communications and higher capacity in Vehicle-to- infra-structure compared to random resource allocation methods and Dynamic proximity aware resource allocation by Ashraf et al. [57]. Also, in the proposed method, the computational complexity of deep neural networks reduces by doubling the weight of the network.

Liu et al. [14], examined baseline round-robin, DRL (DRL, to solve the general edge problem) methods and the proposed hierarchical framework (i.e., a framework consisting of a global tier for allocating virtual machine resources to servers and a local tier for Distributed power management for local servers) in the cloud computing environment to address resource allocation and power management issues. They used a self-cryptographic neural network and a weight-sharing scheme to accelerate convergence speed and control the High-dimension mode space. Their experiments were also implemented with the three methods mentioned on the actual Google cluster traces [58]. The mentioned proposed hierarchical framework optimizes power/energy consumption significantly compared to the baseline round-robin method, but in terms of delay, there is no significant reduction. In a cluster with 30 server classes containing 95,000 job requests, the proposed hierarchy framework saves 53.97% in power/energy consumption and achieves the best compromise between delay and power / energy consumption in a server cluster. On the other hand, by using the proposed framework, the maximum average delay savings with the same energy consumption was 16.16%, while the maximum average power/energy savings with the same delay was 16.20%.

Liang et al. [9] used Deep reinforcement learning to solve wireless resource allocation problems in the Vehicular wireless network environment. Methods with an observer, objective-oriented unsupervised learning paradigm, and learning accelerated optimization paradigm were examined. The proposed system has been shown in Fig. 5. Deep Neural Network (DNN) learning gets a better result than Weighted Minimum Mean-Squared Error (WMMSE) [59] with a significant reduction in computational complexity for the NP-hard power resource allocation problem [60]. The feed-forward network (FNN) and convolutional neural network (CNN) [61] methods in the linear sum assignment programming (LSAP) problem can be used as a real-time solution. The performance of the two unsupervised methods in [62] using DNN is better than the heuristic WMMSE. The Q-learning, REINFORCE, and DDPG algorithms were used for reinforced learning training and these algorithms performed better than WMMSE.

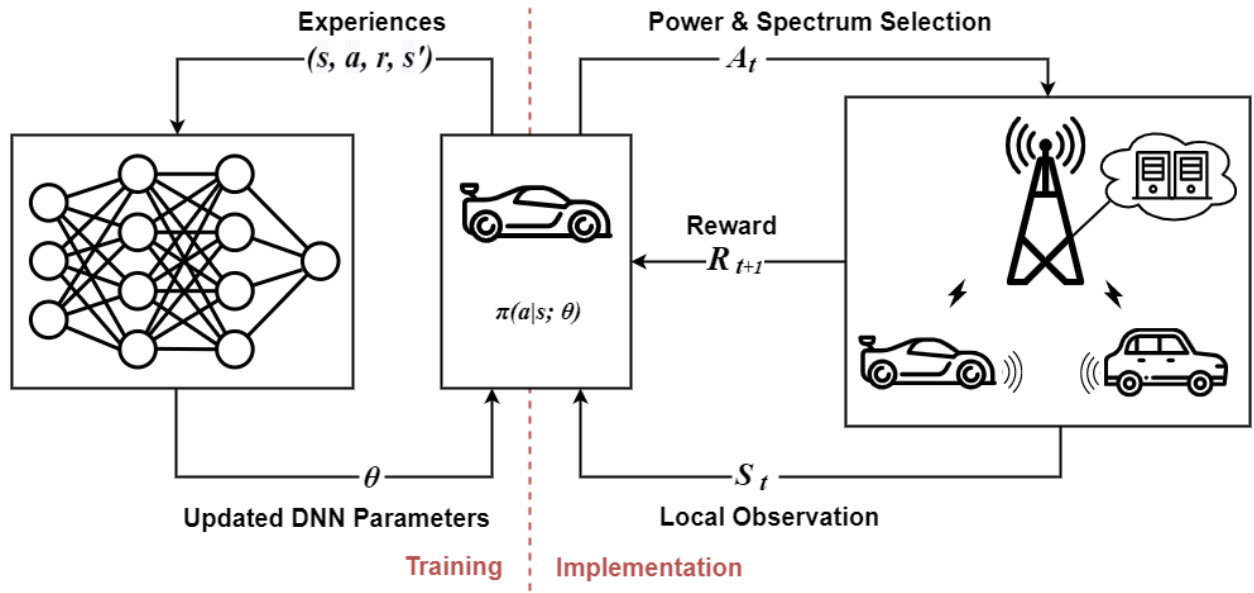

**Figure 6.** Deep reinforcement learning training model for resource allocation in vehicular networks.

Based on Fig.6, each V2V agent makes a local observation of the environment and then utilizes its local copy of the trained Deep reinforcement learning to monitor its resource block selection and power control in a distributed way.

Zhao et al. [63], used distributed deep reinforcement learning for computational resources management, resource allocation, and system complexity reduction in Vehicular Fog Computing environments. They proposed a contract-based incentive mechanism for the allocation of resources in the vehicular fog network. As the number of vehicles increases, the proposed method encourages more vehicles to participate, thus the participation of vehicles increase, Performance quality and efficiency of the system will be maintained well. While in the conventional offloading method, when the vehicle does not cooperate, the computational load returns to the Roadside Unit (RSU) and puts pressure on the RSU.

Zhao et al. [64] used Multi-Agent Deep Reinforcement Learning (MADRL) to reduce co-channel interference, prevent collisions, increasing system power in the proposed smart resource allocation method in train-to-train (T2T) communication. They also used the multi-agent deep Q-network method to solve the train-to-train resource allocation problem, so that the multi-agent deep Q-network method performs better in Successful data transfer, improvement of the throughput of any T2T connection, and improvement of system throughput. To evaluate the proposed multi-agent deep Q-network design, it was compared with the T2T communication resource allocation scheme [65] and the random allocation scheme. In [65], Stackelberg's game theory, for power control and weight factors based on proportional fairness standard, is proposed for channel selection, which (or to solve) solved the resource allocation in the T2T scenario.

Wang et al. [17] used reinforced deep learning in mobile edge computing to allocate a resource called DRLRA (Deep Reinforcement Learning based Resource Allocation). The DRLRA was conducted in Beijing, China, taking into account the actual network topology. As a result, the proposed method improves the average service time as request aggregation districts number increases compared to the OSPF (Open Shortest Path First) method [66].

## 3 Discussion

In this section, at first, the survey articles of the methods are reviewed in Table 2 and the research articles are presented based on ML and DL methods in Table 3 and Table 4 respectively.

**Table 2.** The survey articles related to resource allocation.

| No. | References | No. Citations-Publishers | Deep/Machine learning | Short description | Computing Paradigms |
|---|---|---|---|---|---|
| 1 | Yousefzai et al. [20] | 62-Springer | N/A | Examined the schemes based on cloud computing resources by using effective features | Cloud computing |
| 2 | Atman and Nayan, [21] | 5- Springer | Machine learning | Categorized reinforcement and heuristic learning methods for public safety communications on 5G networks utility | Edge computing |
| 5 | Ghobaei-Arani et al., [24] | 89-Springer | N/A | Six categories of resource management were examined | Fog computing |
| 6 | Hameed et al., [25] | 331- Springer | N/A | Energy efficiency for the resource allocation problem | Cloud computing |
| 7 | Beloglazov et al., [26] | 898-Elsevier | N/A | A discussion on advancements identified in energy-efficient computing | Cloud computing |
| 8 | Shuja et al., [27] | 187-IEEE | N/A | Analyzed mechanisms to control and coordinate data center resources for energy-efficient operations | Cloud computing |
| 9 | Aceto et al., [28] | 724- Elsevier | N/A | Focused on resource monitoring in the cloud computing environment | Cloud computing |
| 10 | Jennings and Stadler, [29] | 543- Springer | N/A | Developed as a framework for resource management in the cloud computing environment | Cloud computing |
| 11 | Goyal and Dadizadeh, [30] | 67-University of British Columbia, Vancouver | N/A | Implementation details of parallel processing frameworks, including Google MapReduce and Microsoft's Dryad | Cloud computing |
| 12 | Hussain et al., [31] | 146- Elsevier | N/A | Presented the work process for commercial cloud computing service providers | Cloud computing |
| 13 | Huang et al., [32] | 66-journal of software | N/A | Studied the Virtual machine migration optimization features | Cloud computing |
| 14 | Ahmed et al., [33] | 84- Springer | N/A | Studied the Virtual machine migration optimization features | Cloud computing |
| 15 | Ahmed et al., [34] | 336-Elsevier | N/A | | |
| 16 | Vinothina et al., [35] | 60-IEEE | N/A | Analyzed the classification of types of strategies and challenges related to resource allocation | Cloud computing |
| 17 | Anuradha and Sumathi, [36] | 56- IEEE | N/A | Resource allocation techniques and strategies in the cloud computing environment were examined | Cloud computing |
| 18 | Mohamaddiah et al., [37] | 45- International Journal of Machine Learning and Computing | N/A | Researched in the field of resource management, In particular, resource allocation and resource monitoring strategies | Cloud computing |
| 19 | RamMohan and Baburaj, [38] | 46- IEEE | N/A | Provided strategies for resource allocation and their applications in the cloud | Cloud computing |

| 20 | Castañeda et al., [39] | 126-IEEE | N/A | Provide a comprehensive overview of the various methodologies used methodologies to approach the aforementioned joint optimization task in the Downlink of MU-MIMO communication systems | Wireless networks |
| 21 | Manvi and Shyam, [40] | 557- Elsevier | N/A | Examined the management methods of resource management approaches | Cloud computing |
| 22 | Su et al., [41] | 47- IEEE | N/A | Examined the techniques and models of resource allocation algorithms in 5G network slicing | 5G telecommunication networks |
| **24** | **In this paper** | --- | **Machine Learning and Deep Learning** | **Reviewing the Machine Learning and Deep Learning methods for resource allocation in different computing paradigms** | **Wireless, 5G, VFC, IoT, Edge, Fog, Cloud** |

Based on Table 2 despite that cloud computing has been received more attention, some problems such as high latency, high jitter, lack of location awareness, limited mobility support, and lack of support for real-time interactions in this environment. At the same time, by emerging IoT and mobile communication in the past few years, articles about edge computing and fog computing, and 5g mobile networks and wireless networks environment are used in the most recent researches. Between the survey articles, we worked on a vast aspect of the computing environment. In addition, in the past few years, the ML and DL methods were used to achieve a better result regarding automatic decision-making for the computing environment in the articles. As can be seen, the most citation is to the article published in Elsevier journal [28] which shows the importance of the problem of resource allocation in the cloud computing environment.

To increase the throughput and performance of a computing environment, it is necessary to use automatic decision-making methods regarding optimal resources allocation using machine learning and deep learning methods. The articles that used ML and DL methods to achieve an optimal resource allocation in different computing paradigms are shown in Table 3 and Table 4 respectively.

**Table 3.** Using ML methods for resource allocation in different Computing Paradigms

| No. | References | No. Citations-Publishers | Short Description | Technique | Computing Paradigms | Language/Platform/Libraries | Results |
|---|---|---|---|---|---|---|---|
| 1 | Shi et al., [42] | 62-IEEE | Proposing MDP method Combined with Bayesian learning to optimize the cost of allocating dynamic resources in the cloud computing for the NFV | MDP-Bayesian Learning | cloud computing | WorkflowSim | Time Ratio: GA/MDP1 708/20=35.4%, Cost Ratio: MDP1/GA 6000/1200=5% |
| 2 | Rohmer et al., [8] | 8-IEEE | Attention to the problem of maximizing the capacity of the P2P streaming system by alternating changing different resource allocation strategies | LB-RA | P2P Streaming System | Python | Mean rejection rate=9.2 %, Max rejection rate= 55.2%, Mean entropy value=6.20, Entropy standard deviation=0.87 |
| 3 | Gai and Qiu, [4] | 102-Elsevier | Use of two RL-based algorithms to create cost mapping tables and optimal resource allocation to obtain high-accuracy QoE of resource allocation | RLMT-RLRA | Internet-of-Things | Java | Average training time: Time/ number of input task= incremental |
| 4 | AlQerm and Shihada, [7] | 27-IEEE | Proposing a resource allocation scheme with embedded online learning algorithm for resource block allocation and 5G network interference control | Online learning | 5G Systems | N/A | Aggregate system throughput= increase, spectral efficiency=increase, Jain's fairness index= increase, Mean SINR= decrease, Average Outage ratio= decrease, Average outage ratio= decrease |
| 5 | Arkian et al., [48] | 75-Springer | Introducing a Q-learning with Fuzzy logic clustering to solve resource constraint problems by grouping vehicles and providing shared resources | Q-Learning with Fuzzy logic clustering | Vehicular Cloud Computing | OMNet++ and SUMO | Cluster head duration= maximum speed changes from 60 to 120 km/h reduced around 15 % for COHORT clustering algorithm, Service discovery delay while increasing the number of vehicles= COHORT/ CROWN= decrease, Service consuming delay while increasing the number of vehicles=COHORT/ CROWN =decrease |
| 6 | Hussin et al., [53] | 17-Wiley Online Library | Use of Q learning algorithm with K-means clustering to perform gap allocation for machine type communication devices in machine to machine communication | Q-learning with K-means clustering | Machine-to-Machine communication | N/A | With increasing the learning rate; The convergence time: decrease, The convergence rate: increase |
| 7 | Salahuddin et al., [19] | 67-IEEE | Proposing reinforcement-learning method based MDP and greedy heuristic methods to minimize overhead in the VCC | RL-based MDP | Vehicular Cloud Computing | Matlab | Cumulative VM migration overhead: minimize |

Based on Table 3 machine learning methods have been used for the resource allocation problem, including MDP-Bayesian learning, LB-RA, RLMT-RLRA, Online learning, Q-learning with Fuzzy logic clustering, Q-learning with K-means clustering, and RL-based MDP. It can be seen that most researchers tried to utilize methods based on reinforcement learning. Studies also demonstrate that the cloud computing environment has been employed more in different periods.

**Table 4.** Using DL methods for resource allocation in different Computing Paradigms

| No. | References | No. Citations-Publishers | Short Description | Technique | Computing Paradigm | Language/Platform/Libraries | Results |
|---|---|---|---|---|---|---|---|
| 1 | Karthiban and Raj, [13] | 9- Springer | present the Deep reinforcement learning algorithm for fair resource allocation to achieve a better resource allocation model in the CC | DRL | Cloud Computing | CloudSim | Average response time= decrease, Average waiting time= decrease, Efficiency= 94% |
| 2 | Chen et al., [55] | 10-IEEE | Reduce the PRT and Average delay for non-safety using DRL-based resource management algorithm in Vehicular Fog Computing | DRL-based Resource management algorithm | Vehicular Fog Computing | N/A | PRT= decrease, Average delay for non-safety = low |
| 3 | Ye et al., [56] | 253- IEEE | Use of a new decentralized resource allocation mechanism for V2V communications applying RDL method | DRL | Vehicle-to-Vehicle Communications | N/A | V2I Capacity= more capacity, V2V Latency= optimal |
| 4 | Liu et al., [14] | 147- IEEE | Provide a new hierarchical framework using DRL combined with LSTM for solving total resource allocation and power management in cloud computing. | DRL-LSTM | Cloud Computing | N/A | Energy usage= low, power/energy=seave to 16.12%, Latency= low to16.67%, |
| 5 | Liang et al., [9] | 73-IEEE | An overview of applying the burgeoning deep learning technology to wireless resource allocation with application to vehicular networks | DRL | Vehicular Networks | N/A | Accuracy: Hungarian algorithm (100%), CNN (92.76%) Classifier accuracy: Graph embedding (83.88%) for 1500 training samples. |
| 6 | Zhao et al., [63] | 12-IEEE | Use the DRL method to implement the proposed contract-based resource management and task offloading scheme to optimize resource management policy and the decision of tasks offloading and improves system performance | Distributed DRL with Adam optimizer | Vehicle Fog Computing | Python | Reduce system complexity, improve computing power and the performance of the entire system. |

| 7 | Zhao et al., [64] | 3-IEEE | Effectively reduce the interference in the system, improve the throughput of T2T links and the system, and ensure the successful transmission probability of the T2T links within the specified time | Multi-Agent DRL | Train to Train | N/A | Improve the throughput of the T2T link, Reduce the co-channel interference in the system effectively |
|---|---|---|---|---|---|---|---|
| 8 | Wang et al., [17] | 129-IEEE | Allocate computing and network resources adaptively, reduce the average service time and balance the use of resources under varying Mobile Edge Computing paradigms by DRL-based Resource Allocation scheme | DRLRA | Mobile Edge Computing | Python | As the minibatch size increases, the convergence speed of the DRLRA algorithm becomes faster |

Although RL-based methods can be effective for optimal resource allocation, deep learning methods are more powerful for cloud-based systems that require high speed, high throughput, and lower delay. The deep learning-based methods in Table 4 are DRL, DRL-LSTM, distributed DRL with Adam optimizer, multi-agent DRL, and DRLRA. The environments used in this table include cloud computing, vehicular fog computing, vehicle to vehicle communications, cloud computing, vehicular networks, vehicle fog computing, train to train, and mobile edge computing. The proposed methods show that DRL-based resource allocation is the best method with the best performance in terms of co-channel interference, system efficiency, energy efficiency, latency, response time, and complexity.

Additionally, for resource allocation problems in different computing paradigms, Figure 7 demonstrates the annual distribution of 22 papers based on a polynomial trend line from 2009 to 2019. Most of the researchers had been studied the resource allocation problem in the cloud computing paradigm from 2009 to 2018. In contrast, the number of three papers have been published in new computing paradigms like edge computing, fog computing, and 5G networks in 2019. Hence the era of cloud computing is passed and we encounter new environment paradigms.

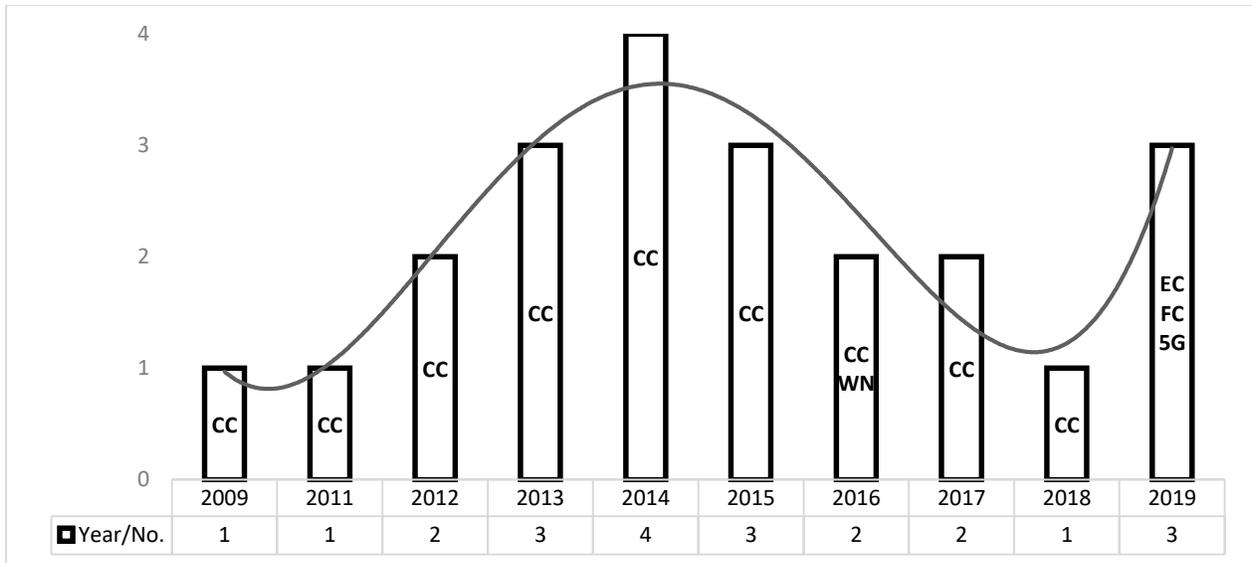

**Figure 7.** The number of survey articles for resource allocation problems in different computing paradigms.

Figure 8 shows the annual distribution of 7 papers based on a polynomial trend line from 2015 to 2018. In 2015, the number of papers has increased, but during the years 2016 to 2018, the number of articles has decreased.

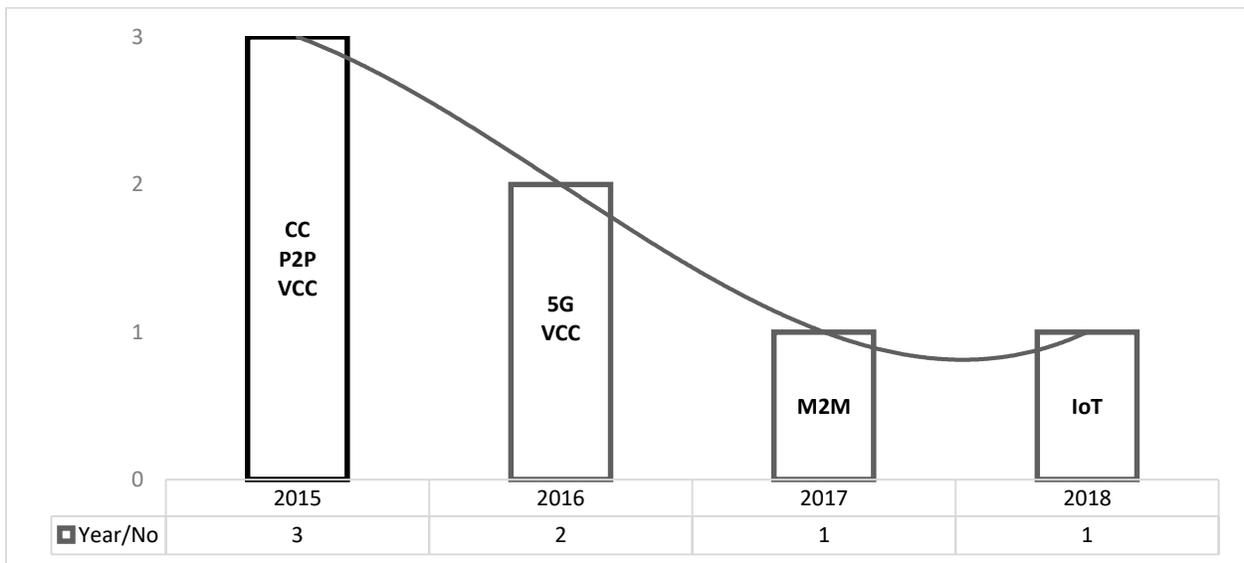

**Figure 8.** The number of research articles regarding ML methods for resource allocation problems in different computing paradigms.

Based on Figure 9, the annual distribution of 7 papers has been illustrated based on a polynomial trend line from 2017 to 2020. From 2017 to 2019, the number of papers has increased, but during the years 2019 and 2020, the number of articles has decreased.

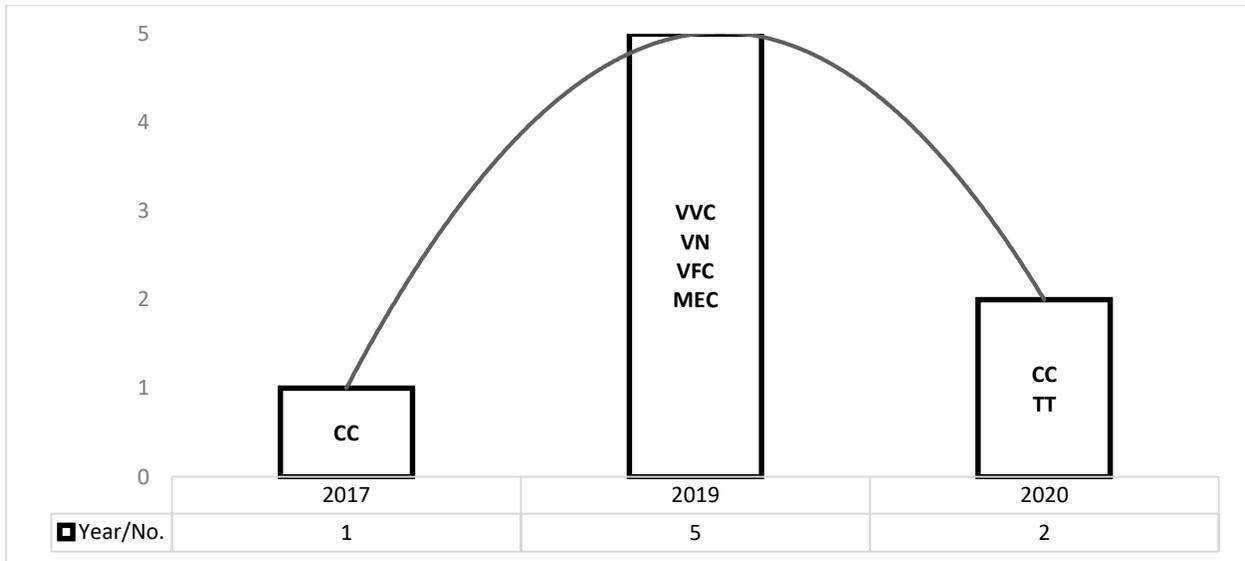

**Figure 9.** The number of research articles regarding DL methods for resource allocation problems in different computing paradigms.

## 4 Open Research Challenges

As mentioned, today machine learning methods are widely used to solve the problem of scheduling tasks and allocating resources in large computing environments such as clouds. Predicting the execution time of programs [67, 68], predicting the success of executing a task [69, 70], modeling the behavior of virtual machines [71], modeling and predicting the amount of energy consumption of a resource [72] are the increasing applications of machine learning methods in better management of the resources of new large computing systems.

Due to the fact that many factors (such as the chance of resource failure and proper estimation of system load) affect the success of a resource management method in large computing systems, resource management is always known as a very complex issue. However, with the development of hardware technologies, today it is possible to monitor and record information related to these systems, and one can have extensive log files of these systems. These conditions, taken together, have provided the conditions for using machine learning methods in the management of these systems, including solving the problem of resource management.

With the emergence of systems such as clouds and the emergence of concepts such as the Internet of Things, we encounter environments that have higher dynamics and their proper management depends on continuous feedback from the environment and choosing the right strategy based on the current conditions of the system. Therefore, in past works, we have seen more use of RL methods in resource management. Since there are many parameters involved in making decisions in these systems and we see these systems becoming increasingly complex and large, it can be expected that in the future, the use of deep learning methods for the modelling the behavior of these systems play a more colorful role.

In this regard, for the better use of machine learning in the management of large systems, there are some challenges that can be mentioned here. First of all, work scheduling and resource management should be done with little time overhead, and therefore the resulting learning model should not be time-consuming in the inference phase. Since the dynamics of new systems necessitates re-learning and continuous learning, it may be better to do the learning phase offline.

The second challenge is concern about data quality. Any disturbance or noise in the data can damage the quality of learning and then the management of the system. Therefore, the important role of monitoring and information recording systems can be seen here. The third challenge is the failure of these systems. These systems are very susceptible to failure and despite many efforts, the failure rate in these systems is still relatively high and it is sometimes difficult to predict these failures. These failures can cause the recording of information needed for system management to be disrupted. Providing alternative and backup methods for situations where the system fails can also be an important challenge.

## 5  Conclusion and future work

By establishing a lot of communication between different systems, including smart devices, storage, servers, communication networks, IoT networks, and the advanced Internet bones and the undeniable impact of these facilities in people's daily lives, optimizing and increasing efficiency is a sensitive issue to be considered. In this context, there are various aspects of sensitivity such as devices, network, and data type that are used collectively, among which resource allocation becomes one of the main bottlenecks in all these aspects. To overcome the challenge of resource allocation, scientists are using innovative resource allocation methods based on new methods of artificial intelligence that can optimize the algorithm according to the data flow during network operation. These measures have moved the industry towards automated resource management on a large and complex scale. In this article, we review and summarize the artificial intelligence (ML and DL) methods to solve the resource allocation problem in terms of response time, energy efficiency, throughput, the performance of the system, cost, service consuming delay, convergence time, latency, etc. According to the results of related research, the deep reinforcement learning method becomes useful for optimal resource allocation. In addition, a comprehensive comparison between these methods has been done in different environments. New methods of resource allocation have become increasingly popular, and computing environments have moved from cloud to fog and edge. While in the last decade, resource allocation in the cloud environment was highly regarded, but resource allocation at the level of smart devices is taken into consideration in the past few years. With the increasing number of smart devices, such as self-driving cars, smart home appliances, smartphones, and robots, which increase the processing power of the edge-computing environment, to increase the quality of service for users it is expected that more research, will be done about resource allocation in this environment in the coming years. In addition, deep reinforcement learning and convolutional neural network algorithms can be proposed combined with meta-heuristic algorithms to find the best model with high scalability and maximum convergence in terms of time delay, speed, and the ability to find global optimization.

## References

[1] Y. Ning, X. Chen, Z. Wang, and X. Li, "An uncertain multi-objective programming model for machine scheduling problem," *International Journal of Machine Learning and Cybernetics,* vol. 8, no. 5, pp. 1493-1500, 2017.
[2] K. Gai, M. Qiu, and X. Sun, "A survey on FinTech," *Journal of Network and Computer Applications,* vol. 103, pp. 262-273, 2018.
[3] R. Liu, C. Vellaithurai, S. S. Biswas, T. T. Gamage, and A. K. Srivastava, "Analyzing the cyber-physical impact of cyber events on the power grid," *IEEE Transactions on Smart Grid,* vol. 6, no. 5, pp. 2444-2453, 2015.
[4] K. Gai and M. Qiu, "Optimal resource allocation using reinforcement learning for IoT content-centric services," *Applied Soft Computing,* vol. 70, pp. 12-21, 2018.
[5] V. Hahanov, *Cyber physical computing for IoT-driven services*. Springer, 2018.
[6] S. Shamshirband *et al.*, "Game theory and evolutionary optimization approaches applied to resource allocation problems in computing environments: A survey," *Mathematical Biosciences and Engineering,* vol. 18, no. 6, pp. 9190-9232, 2021.
[7] I. AlQerm and B. Shihada, "A cooperative online learning scheme for resource allocation in 5G systems," in *2016 IEEE International Conference on Communications (ICC)*, 2016: IEEE, pp. 1-7.
[8] T. Rohmer, A. Nakib, and A. Nafaa, "A learning-based resource allocation approach for P2P streaming systems," *IEEE Network,* vol. 29, no. 1, pp. 4-11, 2015.
[9] L. Liang, H. Ye, G. Yu, and G. Y. Li, "Deep-learning-based wireless resource allocation with application to vehicular networks," *Proceedings of the IEEE,* vol. 108, no. 2, pp. 341-356, 2019.
[10] L. Tong, Y. Li, and W. Gao, "A hierarchical edge cloud architecture for mobile computing," in *IEEE INFOCOM 2016-The 35th Annual IEEE International Conference on Computer Communications*, 2016: IEEE, pp. 1-9.
[11] S. Bitam and A. Mellouk, "Its-cloud: Cloud computing for intelligent transportation system," in *2012 IEEE global communications conference (GLOBECOM)*, 2012: IEEE, pp. 2054-2059.
[12] S. Bitam, A. Mellouk, and S. Zeadally, "VANET-cloud: a generic cloud computing model for vehicular Ad Hoc networks," *IEEE Wireless Communications,* vol. 22, no. 1, pp. 96-102, 2015.
[13] K. Karthiban and J. S. Raj, "An efficient green computing fair resource allocation in cloud computing using modified deep reinforcement learning algorithm," *Soft Computing,* pp. 1-10, 2020.
[14] N. Liu *et al.*, "A hierarchical framework of cloud resource allocation and power management using deep reinforcement learning," in *2017 IEEE 37th International Conference on Distributed Computing Systems (ICDCS)*, 2017: IEEE, pp. 372-382.
[15] I. Stojmenovic and S. Wen, "The fog computing paradigm: Scenarios and security issues," in *2014 federated conference on computer science and information systems*, 2014: IEEE, pp. 1-8.
[16] R. A. Sadek, "Hybrid energy aware clustered protocol for IoT heterogeneous network," *Future Computing and Informatics Journal,* vol. 3, no. 2, pp. 166-177, 2018.
[17] J. Wang, L. Zhao, J. Liu, and N. Kato, "Smart resource allocation for mobile edge computing: A deep reinforcement learning approach," *IEEE Transactions on emerging topics in computing,* 2019.
[18] X. Xu *et al.*, "Dynamic resource allocation for load balancing in fog environment," *Wireless Communications and Mobile Computing,* vol. 2018, 2018.
[19] M. A. Salahuddin, A. Al-Fuqaha, and M. Guizani, "Reinforcement learning for resource provisioning in the vehicular cloud," *IEEE Wireless Communications,* vol. 23, no. 4, pp. 128-135, 2016.
[20] A. Yousafzai *et al.*, "Cloud resource allocation schemes: review, taxonomy, and opportunities," *Knowledge and Information Systems,* vol. 50, no. 2, pp. 347-381, 2017.
[21] A. Othman and N. A. Nayan, "Efficient admission control and resource allocation mechanisms for public safety communications over 5G network slice," *Telecommunication Systems,* vol. 72, no. 4, pp. 595-607, 2019.
[22] B. Han, J. Lianghai, and H. D. Schotten, "Slice as an evolutionary service: Genetic optimization for inter-slice resource management in 5G networks," *IEEE Access,* vol. 6, pp. 33137-33147, 2018.
[23] D. Bega, M. Gramaglia, A. Banchs, V. Sciancalepore, K. Samdanis, and X. Costa-Perez, "Optimising 5G infrastructure markets: The business of network slicing," in *IEEE INFOCOM 2017-IEEE Conference on Computer Communications*, 2017: IEEE, pp. 1-9.
[24] M. Ghobaei-Arani, A. Souri, and A. A. Rahmanian, "Resource management approaches in fog computing: A comprehensive review," *Journal of Grid Computing,* pp. 1-42, 2019.
[25] A. Hameed *et al.*, "A survey and taxonomy on energy efficient resource allocation techniques for cloud computing systems," *Computing,* vol. 98, no. 7, pp. 751-774, 2016.
[26] A. Beloglazov, R. Buyya, Y. C. Lee, and A. Zomaya, "A taxonomy and survey of energy-efficient data centers and cloud computing systems," in *Advances in computers*, vol. 82: Elsevier, 2011, pp. 47-111.
[27] J. Shuja *et al.*, "Survey of techniques and architectures for designing energy-efficient data centers," *IEEE Systems Journal,* vol. 10, no. 2, pp. 507-519, 2014.

[28] G. Aceto, A. Botta, W. De Donato, and A. Pescapè, "Cloud monitoring: A survey," *Computer Networks,* vol. 57, no. 9, pp. 2093-2115, 2013.
[29] B. Jennings and R. Stadler, "Resource management in clouds: Survey and research challenges," *Journal of Network and Systems Management,* vol. 23, no. 3, pp. 567-619, 2015.
[30] A. Goyal and S. Dadizadeh, "A survey on cloud computing," *University of British Columbia Technical Report for CS,* vol. 508, pp. 55-58, 2009.
[31] H. Hussain *et al.*, "A survey on resource allocation in high performance distributed computing systems," *Parallel Computing,* vol. 39, no. 11, pp. 709-736, 2013.
[32] L. Huang, H.-s. Chen, and T.-t. Hu, "Survey on Resource Allocation Policy and Job Scheduling Algorithms of Cloud Computing1," *JSW,* vol. 8, no. 2, pp. 480-487, 2013.
[33] R. W. Ahmad, A. Gani, S. H. A. Hamid, M. Shiraz, F. Xia, and S. A. Madani, "Virtual machine migration in cloud data centers: a review, taxonomy, and open research issues," *The Journal of Supercomputing,* vol. 71, no. 7, pp. 2473-2515, 2015.
[34] R. W. Ahmad, A. Gani, S. H. A. Hamid, M. Shiraz, A. Yousafzai, and F. Xia, "A survey on virtual machine migration and server consolidation frameworks for cloud data centers," *Journal of network and computer applications,* vol. 52, pp. 11-25, 2015.
[35] V. Vinothina, R. Sridaran, and P. Ganapathi, "A survey on resource allocation strategies in cloud computing," *International Journal of Advanced Computer Science and Applications,* vol. 3, no. 6, pp. 97-104, 2012.
[36] V. Anuradha and D. Sumathi, "A survey on resource allocation strategies in cloud computing," in *International Conference on Information Communication and Embedded Systems (ICICES2014)*, 2014: IEEE, pp. 1-7.
[37] M. H. Mohamaddiah, A. Abdullah, S. Subramaniam, and M. Hussin, "A survey on resource allocation and monitoring in cloud computing," *International Journal of Machine Learning and Computing,* vol. 4, no. 1, pp. 31-38, 2014.
[38] N. R. Mohan and E. B. Raj, "Resource Allocation Techniques in Cloud Computing--Research Challenges for Applications," in *2012 fourth international conference on computational intelligence and communication networks*, 2012: IEEE, pp. 556-560.
[39] E. Castaneda, A. Silva, A. Gameiro, and M. Kountouris, "An overview on resource allocation techniques for multi-user MIMO systems," *IEEE Communications Surveys & Tutorials,* vol. 19, no. 1, pp. 239-284, 2016.
[40] S. S. Manvi and G. K. Shyam, "Resource management for Infrastructure as a Service (IaaS) in cloud computing: A survey," *Journal of network and computer applications,* vol. 41, pp. 424-440, 2014.
[41] R. Su *et al.*, "Resource allocation for network slicing in 5G telecommunication networks: A survey of principles and models," *IEEE Network,* vol. 33, no. 6, pp. 172-179, 2019.
[42] R. Shi *et al.*, "MDP and machine learning-based cost-optimization of dynamic resource allocation for network function virtualization," in *2015 IEEE International Conference on Services Computing*, 2015: IEEE, pp. 65-73.
[43] N. M. K. Chowdhury and R. Boutaba, "A survey of network virtualization," *Computer Networks,* vol. 54, no. 5, pp. 862-876, 2010.
[44] X. Cheng, C. Dale, and J. Liu, "Statistics and social network of youtube videos," in *2008 16th Interntional Workshop on Quality of Service*, 2008: IEEE, pp. 229-238.
[45] T. R. Omar, A. E. Kamal, and J. M. Chang, "Downlink spectrum allocation in 5g hetnets," in *2014 International Wireless Communications and Mobile Computing Conference (IWCMC)*, 2014: IEEE, pp. 12-17.
[46] S. Rostami, K. Arshad, and P. Rapajic, "A joint resource allocation and link adaptation algorithm with carrier aggregation for 5G LTE-Advanced network," in *2015 22nd International Conference on Telecommunications (ICT)*, 2015: IEEE, pp. 102-106.
[47] S. A. Kazmi *et al.*, "Resource management in dense heterogeneous networks," in *2015 17th Asia-Pacific Network Operations and Management Symposium (APNOMS)*, 2015: IEEE, pp. 440-443.
[48] H. R. Arkian, R. E. Atani, A. Diyanat, and A. Pourkhalili, "A cluster-based vehicular cloud architecture with learning-based resource management," *The Journal of Supercomputing,* vol. 71, no. 4, pp. 1401-1426, 2015.
[49] H. R. Arkian, R. E. Atani, and S. Kamali, "FcVcA: A fuzzy clustering-based vehicular cloud architecture," in *2014 7th International Workshop on Communication Technologies for Vehicles (Nets4Cars-Fall)*, 2014: IEEE, pp. 24-28.
[50] I. Tal and G.-M. Muntean, "User-oriented fuzzy logic-based clustering scheme for vehicular ad-hoc networks," in *2013 IEEE 77th Vehicular Technology Conference (VTC Spring)*, 2013: IEEE, pp. 1-5.
[51] M. Gerla and J. T.-C. Tsai, "Multicluster, mobile, multimedia radio network," *Wireless networks,* vol. 1, no. 3, pp. 255-265, 1995.
[52] K. Mershad and H. Artail, "Finding a STAR in a Vehicular Cloud," *IEEE Intelligent transportation systems magazine,* vol. 5, no. 2, pp. 55-68, 2013.

[53] F. Hussain, A. Anpalagan, A. S. Khwaja, and M. Naeem, "Resource allocation and congestion control in clustered M2M communication using Q-learning," *Transactions on Emerging Telecommunications Technologies,* vol. 28, no. 4, p. e3039, 2017.

[54] M. A. Salahuddin, A. Al-Fuqaha, and M. Guizani, "Software-defined networking for rsu clouds in support of the internet of vehicles," *IEEE Internet of Things journal,* vol. 2, no. 2, pp. 133-144, 2014.

[55] X. Chen, S. Leng, K. Zhang, and K. Xiong, "A machine-learning based time constrained resource allocation scheme for vehicular fog computing," *China Communications,* vol. 16, no. 11, pp. 29-41, 2019.

[56] H. Ye, G. Y. Li, and B.-H. F. Juang, "Deep reinforcement learning based resource allocation for V2V communications," *IEEE Transactions on Vehicular Technology,* vol. 68, no. 4, pp. 3163-3173, 2019.

[57] M. I. Ashraf, M. Bennis, C. Perfecto, and W. Saad, "Dynamic proximity-aware resource allocation in vehicle-to-vehicle (V2V) communications," in *2016 IEEE Globecom Workshops (GC Wkshps)*, 2016: IEEE, pp. 1-6.

[58] C. Reiss, J. Wilkes, and J. L. Hellerstein, "Google cluster-usage traces: format+ schema," *Google Inc., White Paper,* pp. 1-14, 2011.

[59] Y. S. Nasir and D. Guo, "Multi-agent deep reinforcement learning for dynamic power allocation in wireless networks," *IEEE Journal on Selected Areas in Communications,* vol. 37, no. 10, pp. 2239-2250, 2019.

[60] H. Sun, X. Chen, Q. Shi, M. Hong, X. Fu, and N. D. Sidiropoulos, "Learning to optimize: Training deep neural networks for interference management," *IEEE Transactions on Signal Processing,* vol. 66, no. 20, pp. 5438-5453, 2018.

[61] W. Lee, M. Kim, and D.-H. Cho, "Deep power control: Transmit power control scheme based on convolutional neural network," *IEEE Communications Letters,* vol. 22, no. 6, pp. 1276-1279, 2018.

[62] Q. Shi, M. Razaviyayn, Z.-Q. Luo, and C. He, "An iteratively weighted MMSE approach to distributed sum-utility maximization for a MIMO interfering broadcast channel," *IEEE Transactions on Signal Processing,* vol. 59, no. 9, pp. 4331-4340, 2011.

[63] J. Zhao, M. Kong, Q. Li, and X. Sun, "Contract-Based Computing Resource Management via Deep Reinforcement Learning in Vehicular Fog Computing," *IEEE Access,* vol. 8, pp. 3319-3329, 2019.

[64] J. Zhao, Y. Zhang, Y. Nie, and J. Liu, "Intelligent Resource Allocation for Train-to-Train Communication: A Multi-Agent Deep Reinforcement Learning Approach," *IEEE Access,* vol. 8, pp. 8032-8040, 2020.

[65] Q. Zhou, X. Hu, J. Lin, and Z. Wu, "Train-to-train communication resource allocation scheme for train control system," in *2018 10th International Conference on Communication Software and Networks (ICCSN)*, 2018: IEEE, pp. 210-214.

[66] M. Caria, T. Das, A. Jukan, and M. Hoffmann, "Divide and conquer: Partitioning OSPF networks with SDN," in *2015 IFIP/IEEE International Symposium on Integrated Network Management (IM)*, 2015: IEEE, pp. 467-474.

[67] T.-P. Pham, J. J. Durillo, and T. Fahringer, "Predicting workflow task execution time in the cloud using a two-stage machine learning approach," *IEEE Transactions on Cloud Computing,* vol. 8, no. 1, pp. 256-268, 2017.

[68] I. Pietri, G. Juve, E. Deelman, and R. Sakellariou, "A performance model to estimate execution time of scientific workflows on the cloud," in *2014 9th Workshop on Workflows in Support of Large-Scale Science*, 2014: IEEE, pp. 11-19.

[69] E. Shirzad and H. Saadatfar, "Job failure prediction in Hadoop based on log file analysis," *International Journal of Computers and Applications,* vol. 44, no. 3, pp. 260-269, 2022.

[70] H. Saadatfar, H. Fadishei, and H. Deldari, "Predicting job failures in AuverGrid based on workload log analysis," *New Generation Computing,* vol. 30, no. 1, pp. 73-94, 2012.

[71] S. S. Gill *et al.*, "ThermoSim: Deep learning based framework for modeling and simulation of thermal-aware resource management for cloud computing environments," *Journal of Systems and Software,* vol. 166, p. 110596, 2020.

[72] M. K. M. Shapi, N. A. Ramli, and L. J. Awalin, "Energy consumption prediction by using machine learning for smart building: Case study in Malaysia," *Developments in the Built Environment,* vol. 5, p. 100037, 2021.